# Enhancing Medical Data Analysis through AI-Enhanced Locally Linear Embedding: Applications in Medical Point Location and Imagery

[1] Hassan Khalid
School of Electrical Engineering and Computer Science
National University of Science and Technology
Islamabad, Pakistan
hkhalid.msai23seecs@seecs.edu.pk

[2] Muhammad Mahad Khaliq
School of Electrical Engineering and Computer Science (SEECS)
National University of Science and Technology, NUST
Islamabad, Pakistan
mkhaliq.msee21seecs@seecs.edu.pk

[3] Muhammad Jawad Bashir
School of Electrical Engineering and Computer Science (SEECS)
National University of Science and Technology, NUST
Islamabad, Pakistan
mbashir.msds20seecs@seecs.edu.pk

*Abstract*—The rapid evolution of Artificial intelligence in healthcare has opened avenues for enhancing various processes, including medical billing and transcription. This paper introduces an innovative approach by integrating AI with Locally Linear Embedding (LLE) to revolutionize the handling of high-dimensional medical data. This AI-enhanced LLE model is specifically tailored to improve the accuracy and efficiency of medical billing systems and transcription services. By automating these processes, the model aims to reduce human error and streamline operations, thereby facilitating faster and more accurate patient care documentation and financial transactions. This paper provides a comprehensive mathematical model of AI-enhanced LLE, demonstrating its application in real-world healthcare scenarios through a series of experiments. The results indicate a significant improvement in data processing accuracy and operational efficiency. This study not only underscores the potential of AI-enhanced LLE in medical data analysis but also sets a foundation for future research into broader healthcare applications.

## I. Introduction

Locally Linear Embedding (LLE), augmented with AI, is a transformative technique for simplifying and interpreting complex high-dimensional medical data. This AI-enhanced LLE approach does not merely manage the intricacies of raw medical data; instead, it preserves crucial relationships between data points, such as patient information, billing codes, and transcription records, when transitioned into a lower-dimensional space. By deconstructing each data point into a weighted sum of its nearest neighbors, the technique efficiently finds a simpler, yet accurate, representation. This representation not only maintains the essential connections within the data but also enhances the precision and efficiency of medical billing and transcription processes, making it a valuable tool in healthcare data management.

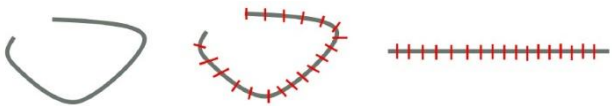

Fig. 1. Piece-wise local unfolding of manifold by LLE (in this example from two dimensions to one intrinsic dimension). This local unfolding is expected to totally unfold the manifold properly. [3]

The Locally Linear Embedding (LLE) algorithm and its extensions, such as the Spherical Locally Linear Embedding (sLLE), address challenges related to data analysis and visualization, making them relevant to several UN sustainability goals. LLE's ability to simplify complex datasets and uncover hidden structures can contribute to Goal 9 (Industry, Innovation, and Infrastructure) by enhancing data analysis techniques for sustainable infrastructure development. Additionally, the application of LLE in seismic attribute extraction and MRI-based Alzheimer's disease classification aligns with Goal 3 (Good Health and Well-being) by improving diagnostic capabilities and healthcare outcomes. Furthermore, the utilization of sLLE in tomography supports Goal 11 (Sustainable Cities and Communities) by aiding in urban planning and resource management. Overall, the mathematical model and applications of LLE offer valuable insights for achieving various UN sustainability goals through enhanced data analysis and decision-making processes.

Breaking down LLE's mathematical steps into simpler terms, the algorithm first identifies the closest neighbors for each data point using basic distance calculations. Then, it calculates weights that minimize the difference between the original data point and its simplified version based on the chosen neighbors. Finally, LLE transforms the data into a simpler form by finding a representation in fewer dimensions that preserves the basic structure captured by these weights. Widely used in areas like image processing, pattern recognition, and data visualization, LLE proves valuable for making sense of complicated datasets. Its knack for revealing hidden patterns and making data more straightforward for interpretation and visualization highlights its importance in various applications. Figure-1 elaborates how a general LLE model works.

## II. Problem Overview

In this paper, the primary focus is on the Locally Linear Embedding (LLE) algorithm, augmented with artificial intelligence (AI), and its applications in medical billing and transcription. LLE, a renowned nonlinear dimensionality reduction technique, is crucial for preserving the local geometry of high-dimensional medical data in a more manageable lower-dimensional space. Despite its effectiveness, LLE's sensitivity to noise, outliers, and neighborhood selection can pose challenges in the precise environments of medical data handling. To address these issues, we explore various AI-driven enhancements to LLE, including the adoption of alternative distance metrics, optimized neighborhood selection rules, and the integration of advanced processing steps. These modifications aim to bolster the performance, stability, and accuracy of LLE, thereby enhancing its applicability in complex medical data analysis tasks, particularly in improving the reliability and efficiency of medical billing and transcription processes.

### III. RELATED WORK

The review paper [1] provides a comprehensive overview of the advancements in the field of nonlinear dimensionality reduction, specifically focusing on the LLE algorithm and its extensions. The authors meticulously review various techniques proposed in recent years to address the limitations of LLE, offering valuable insights into the challenges posed by noisy, sparse, or unevenly distributed datasets. The paper covers a wide array of LLE variants, discussing their mathematical formulations, strengths, and applications, providing a solid foundation for researchers and practitioners interested in manifold learning, dimensionality reduction, and pattern recognition. It serves as an authoritative reference for those seeking a deeper understanding of LLE and its evolving landscape.

In the publication on Nonlinear Dimensionality Reduction by LLE [2], the authors introduce a novel unsupervised learning algorithm known as Locally Linear Embedding (LLE). The authors address the critical challenge of dimensionality reduction in exploratory data analysis and visualization. LLE computes low-dimensional embeddings of high-dimensional inputs by preserving local neighborhoods, providing a global coordinate system without encountering local minima. The method demonstrates its efficacy in capturing the global structure of nonlinear manifolds, such as those found in images of faces or text documents. The paper emphasizes the importance of understanding the intrinsic geometry of high-dimensional data and highlights LLE's advantages over traditional methods like principal component analysis (PCA) and multidimensional scaling (MDS). The work showcases applications of LLE to face images and word-document counts, revealing meaningful embeddings. With its favorable properties and potential applications in diverse scientific domains, this seminal paper serves as a foundational reference for researchers exploring nonlinear dimensionality reduction techniques.

### IV. MATHEMATICAL MODEL

The mathematical model proposed herein adapts Locally Linear Embedding (LLE) integrated with AI, as a sophisticated dimensional reduction technique for medical data analysis. This model aims to represent each complex data point—such as medical billing entries and transcription texts—in a high-dimensional space as a weighted sum of its K nearest neighbors, thus minimizing the reconstruction error. The AI-enhanced approach also ensures invariance to translation, rotation, and scaling, maintained through a normalization constraint. Employing Lagrange multipliers, this model offers a constrained minimization strategy that yields an exact solution for the weights, ensuring an optimal local representation. This optimized representation not only captures the intrinsic data geometry but also enhances the precision and efficiency of processing medical data. The resulting solution effectively reduces the dimensional complexity of medical data, providing clearer insights into its structured patterns and supporting more accurate data handling in healthcare applications. Consider a data-point $x_i \in \mathbb{R}^D$ where it represents any row from the dataset. The goal is to represent the $x_i$ with dimension P such that P<D.

The first step involves local linearization where each datapoint is to be represented by weighted sum (or linear combination) of its K nearest neighbors. The goal is to represent the datapoint as closely as possible hence the optimization problem thus formulated is represented by

$$\text{minimize} \left\| x_i - \sum_{j \in N_i} \omega_{ij} x_j \right\|^2 \quad (1)$$

Where, $N_i$ represents the nearest neighbors for $i^{th}$ example and $x_j$ the datapoints in neighborhood of $x_i$. This minimization function ensures that the $x_i$ is as closely represented by its neighbor as possible.

$$x_i \approx \sum_{j \in N_i} \omega_{ij} x_j \quad (2)$$

This optimization is subjected to the following constraint.

$$\sum_{j \in N_i} \omega_{ij} = 1 \quad (3)$$

This constraint ensures that the reconstruction is invariant to translation, rotation, and scaling, which is desirable for capturing the intrinsic geometry of the data. In other words, it normalizes the weights for example preserving its local structure in the data. Using Eq. 3, Eq. 1 can be written as

$$\left\| \sum_{j \in N_i} \omega_{ij}(x_i - x_j) \right\|^2 \quad (4)$$

In matrix form, Eq. 3 can be written as

$$\mathbf{1}^T \omega_i = 1 \quad (5)$$

Where $\omega_i \in \mathbb{R}^K$ is vector containing weights for K neighborhood points. Additionally,

$$Z_i = [x_i - x_1 \cdots x_i - x_j \cdots x_i - x_K] \in \mathbb{R}^{D \times K} \quad (6)$$

Hence, Eq. 4 can be written as

$$\|Z_i \omega_i\|^2 = (Z_i \omega_i)^T (Z_i \omega_i)$$
$$\|Z_i \omega_i\|^2 = \omega_i^T Z_i^T Z_i \omega_i$$
$$\|Z_i \omega_i\|^2 = \omega_i^T G_i \omega_i \quad (7)$$

Where $G_i \in \mathbb{R}^{K \times K}$ is a local gram matrix which positive semidefinite with nonnegative eigen values.
Hence final optimization problem is defined as,

$$\text{minimize } \omega_i^T G_i \omega_i \text{ subject to } \mathbf{1}^T \omega_i = 1 \quad (8)$$

This constrained optimization problem can be solved via Lagrange multipliers. Lagrange equation is defined as

$$\mathcal{L} = \omega_i^T G_i \omega_i - \lambda_i (\mathbf{1}^T \omega_i - 1)$$

The minima can be obtained by setting partial derivatives to zero.

$$\mathbb{R}^K \ni \frac{\partial \mathcal{L}}{\partial \omega_i} = 2 G_i \omega_i - \lambda_i \mathbf{1} = \mathbf{0}$$

$$\omega_i = \frac{\lambda_i}{2} G_i^{-1} \mathbf{1} \quad (9)$$

$$\mathbb{R} \ni \frac{\partial \mathcal{L}}{\partial \lambda} = \mathbf{1}^T \boldsymbol{\omega}_i - 1 = 0$$

$$\mathbf{1}^T \boldsymbol{\omega}_i = 1 \quad (10)$$

Putting Eq. 9 in Eq. 10,

$$\frac{\lambda_i}{2} \mathbf{1}^T G_i^{-1} \mathbf{1} = 1$$

$$\lambda_i = \frac{2}{\mathbf{1}^T G_i^{-1} \mathbf{1}} \quad (11)$$

This value for Lagrange multiplier produces $\omega_i$ for which minimum loss is obtained.

$$\boldsymbol{\omega}_i = \frac{G_i^{-1} \mathbf{1}}{\mathbf{1}^T G_i^{-1} \mathbf{1}} \quad (12)$$

Eq. 12 represents the exact solution for $\omega_i$ for datapoint $x_i \in \mathbb{R}^D$ to minimize the optimization function.

After local linearization of all datapoints, next step is to embed each datapoint onto a lower dimension P using the weights determined by Eq. 12. The goal is to find low dimensional coordinates $y_i \in \mathbb{R}^P$ against each $x_i$. $y_i$ is represented in the same manner as $x_i$ i.e. weighted sum or linear combination of the nearest neighbors $y_j$.

$$\boldsymbol{y}_i \approx \sum_{j \in N_i} \omega_{ij} \boldsymbol{y}_j \quad (13)$$

The weights determined in previous step can be represented in the form of a weight matrix $W \in \mathbb{R}^{n \times n}$, where n is number of datapoints in the data, which contains the weights for all data points.

$$\mathbb{R}^{n \times n} \ni W = \begin{bmatrix} \omega_{11} & \cdots & \omega_{1n} \\ \vdots & \ddots & \vdots \\ \omega_{n1} & \cdots & \omega_{nn} \end{bmatrix} \quad (14)$$

where
$$\omega_{ij} = \begin{cases} 0 & j = i \\ \omega_{ij} & j \in N_i \\ 0 & j \notin N_i \end{cases}$$

For a given row i in weight matrix, all entries will be zero except the columns corresponding to K nearest neighbors of $i^{th}$ example. Hence, by definition

$$\boldsymbol{W1} = \mathbf{1} \quad (15)$$

Similarly, a matrix $Y \in \mathbb{R}^{n \times P}$ can be defined representing the original data in P dimension.

$$\mathbb{R}^{n \times P} \ni Y = \begin{bmatrix} \boldsymbol{y}_1^T \\ \vdots \\ \boldsymbol{y}_n^T \end{bmatrix} \quad (16)$$

Referring to Eq. 13, the optimization function for lower dimensional embedding can be defined as

$$\text{minimize } \Phi(Y) = \sum_{i=1}^{n} \left\| \boldsymbol{y}_i - \sum_{j \in N_i} \omega_{ij} \boldsymbol{y}_j \right\|^2 \quad (17)$$

For simplification, let's assume P = 1 as the solution for general case (P>1) will exhibit similar manner.

$$\Phi(Y) = \sum_{i=1}^{n} \left[ y_i^2 - y_i \left( \sum_{j \in N_i} \omega_{ij} y_j \right) - \left( \sum_{j \in N_i} \omega_{ij} y_j \right) y_i + \left( \sum_{j \in N_i} \omega_{ij} y_j \right)^2 \right]$$

$$\Phi(Y) = Y^T Y - Y^T (WY) - (WY)^T Y + (WY)^T (WY)$$

$$\Phi(Y) = ((I - W)Y)^T ((I - W)Y)$$

$$\Phi(Y) = Y^T (I - W)^T (I - W) Y$$

$$\Phi(Y) = Y^T M Y \quad (18)$$

Where $M \in \mathbb{R}^{n \times n}$ is a positive semidefinite matrix.
The matrix Y, which is to be determined, represents nxP. Constraints can be applied on Y to get a unique solution.
It can be assumed that the coordinates in P dimension are centered around origin.

$$\frac{1}{n} \sum_{i=1}^{n} \boldsymbol{y}_i = \mathbf{0} \quad (19)$$

Additionally, the optimization problem can be converted into eigenvalue problem by following

$$\mathbb{R}^{P \times P} \ni \frac{1}{n} Y^T Y = I \quad (20)$$

Eq. 19 and 20 together ensure that the covariance of matrix Y is an identity matrix. In other words, the columns of matrix Y are orthonormal to each other.
Now using Lagrange multipliers to apply Eq. 20 on objection function,

$$\mathcal{L} = Y^T M Y - \lambda \left( \frac{1}{n} Y^T Y - I \right)$$

But since, it is assumed that P = 1, identity matrix simply becomes 1.

$$\mathcal{L} = Y^T M Y - \lambda \left( \frac{1}{n} Y^T Y - 1 \right)$$

$$\mathbb{R}^{n \times P} \ni \frac{\partial \mathcal{L}}{\partial Y} = 2MY - \frac{2}{n} \lambda Y = 0$$

or $$MY = \frac{\lambda}{n} Y \quad (21)$$

Eq. 21 states that the Y is an eigen vector of matrix M. Since it is a minimization problem, there is a need to select such an eigen value of matrix M which minimizes the loss function (Eq. 18). The eigen vector corresponding to this eigen value will be the Y which will represent the solution of P dimensional embedding of data.
It is previously stated that matrix M is positive semidefinite with non-negative eigen values. The minimum eigen value of M will always be zero with the corresponding eigenvector as column vector of ones. This can be explained by referred to Eq. 15 which translates to following

$$(I - W)\mathbf{1} = \mathbf{0}$$

$$(I - W)^T (I - W) \mathbf{1} = \mathbf{0}$$

$$M\mathbf{1} = \mathbf{0} \quad (22)$$

This will be discarded since it is constant for all problems and doesn't produce a meaningful coordinate for the new

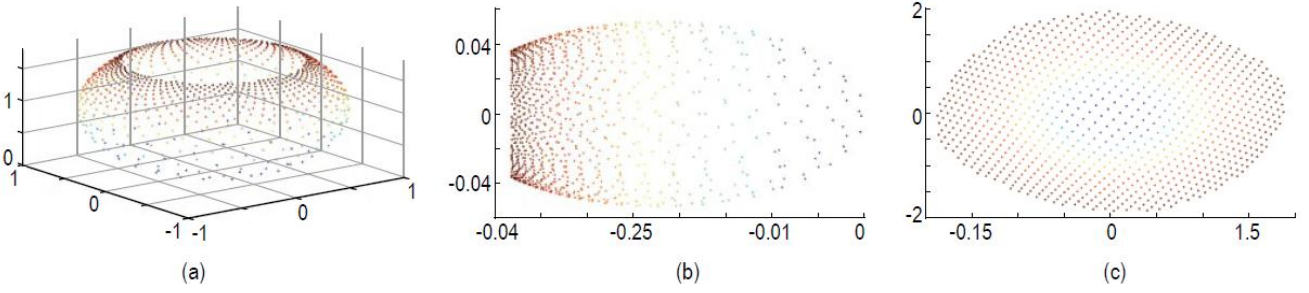

Fig. 2. Comparison of model tests [5]
(a) Punctured sphere samples with N = 800   (b) The PCA dimensionality reduction   (c) The LLE dimensionality reduction with K = 12.

dimension. Therefore, the next smallest eigen value will be selected to produce solution.

The same logic will be translated to the general case where for a given dimension P, P+1 eigenvectors of M will be determined, while discarding the zero eigen-value-vector, to create matrix Y ∈ R$^{nxP}$.

$$\mathbb{R}^{n \times P} \ni Y = [v_{n-1} \quad \cdots \quad v_{n-P}] \quad (23)$$

Where v represents the eigenvectors of matrix M and the subscript n-1 denotes the second smallest eigenvalue going all the way to n-P which denotes (P+1)th smallest eigenvalue.

## V. APPLICATIONS

LLE demonstrates remarkable versatility across diverse fields. Its application in tomography unveils hidden structures in spherical datasets, showcasing its adaptability in complex spatial analyses. In seismic attribute extraction, LLE proves transformative, unraveling intricate geological features and revolutionizing data analysis in geophysics. Furthermore, LLE's role extends into healthcare, particularly in MRI-based Alzheimer's disease classification, emphasizing its potential for enhancing diagnostic capabilities. These varied applications collectively underscore LLE's prowess as a versatile tool for unraveling complex data structures, spanning geophysics to healthcare. Hence, applications are:

### A. Seismic attribute extraction

A pivotal application of Locally Linear Embedding (LLE) in seismic attribute extraction, drawing insights from the notable work by Liu et al [5]. The researchers delve into the challenging terrain of the Tarim basin's carbonate platform, employing LLE to map high-dimensional seismic attribute data into a lower-dimensional space. The comparison with Principal Component Analysis (PCA) reveals LLE's superior ability to preserve intrinsic data structure, especially in capturing complex geological features like reef-shoal reservoirs.

Figure-2 visually encapsulates the dimensionality reduction differences between LLE and PCA, using a punctured sphere model. The illustration compellingly demonstrates LLE's efficacy in retaining coherent data structures, making it a potent tool for extracting intrinsic features from non-linear manifold data.

The insights garnered from this study significantly contribute to the discourse on seismic attribute analysis, emphasizing LLE's potential to enhance reservoir prediction accuracy. When discussing applications of dimensionality reduction techniques, [5] work stands out as a reference, showcasing LLE's prowess in improving seismic attribute analyses and advancing reservoir characterization.

### B. MRI based Alzheimer's disease classification

The application of Locally Linear Embedding (LLE) in the analysis of structural MRI scans for Alzheimer's disease (AD). Using data from the Alzheimer's Disease Neuroimaging Initiative (ADNI), the study focuses on the potential of LLE as a nonlinear dimensionality reduction technique to reveal intricate patterns associated with AD. The researchers emphasize the importance of robust image processing, and compare LLE with other dimensionality reduction methods. It delves into predictive markers for AD, incorporating cognitive reserve in early diagnosis, and considers the impact of long-term cost-effectiveness of AD treatments. The study contributes to a refined understanding of AD progression, offering valuable insights for early detection. The researchers advocate for data sharing and emphasize the collaborative nature of AD research, aligning with the broader goal of addressing the global burden of AD. Overall, this paper provides a comprehensive exploration of LLE's potential in the context of AD classification, contributing to the ongoing efforts to advance diagnosis and intervention strategies for this complex neurodegenerative disease. [6]

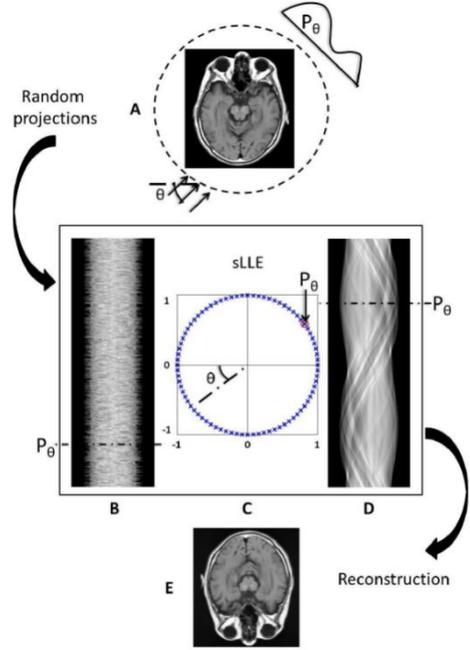

Fig. 3. sLLE Reconstruction Flowchart [7]

### C. Utilization of Spherical Locally Linear Embedding (sLLE) in Tomography

sLLE is an extension of the LLE algorithm. While LLE excels in nonlinear dimensionality reduction, it struggles with embedding points on non-flat manifolds like spheres. sLLE addresses this limitation by introducing a novel approach to constrain the embedding of high-dimensional data on a sphere. This extension enhances the applicability of LLE, allowing it to effectively map data onto a sphere with lower dimensionality to handle scenarios where the underlying data exhibits spherical constraints, making it a valuable tool in various fields, including tomographic image reconstruction. A key strength of this method lies in its ability to handle diverse imaging modalities, including phantom images, MR brain images, and cryoEM density images. The authors [7] present compelling evidence of sLLE's superior performance compared to traditional LLE, particularly in scenarios involving non-symmetric images.

This figure explains that the measurements are taken at various angles, forming a stacked representation (B). The sLLE algorithm is then employed to embed these measurements into a circle (C), associating each point on the circle with a specific projection ($P_\theta$). The circle-based embedding effectively organizes the projections based on their relative view angles. The sorted projection data is presented in (D). Finally, figure (E) displays the reconstructed and refashioned image obtained from the systematized projection data, with the reconstructed image undergoing a global rotation transform compared to the unprecedented image. The figure may depict the reconstructed images alongside their original counterparts, providing a visual testament to the algorithm's success in accurately representing the underlying structures even when viewed from different perspectives.

## Conclusion

In conclusion, this document provides a comprehensive exploration of Locally Linear Embedding (LLE), optimized with artificial intelligence techniques, for applications in medical billing and transcription. The mathematical model of LLE is detailed, highlighting steps such as neighborhood selection and the computation of reconstruction weights. We address the inherent limitations of traditional LLE, such as sensitivity to noise and complex eigen-problems, and introduce AI-driven modifications that improve its robustness and accuracy. This review also synthesizes related work, highlighting advancements by researchers who have expanded the capabilities of LLE to meet these challenges. Through applications in medical data analysis, we demonstrate the effectiveness of LLE in simplifying complex medical datasets, improving the accuracy of billing codes, and enhancing the reliability of transcription processes. Each application illustrates the adaptability and efficacy of LLE in navigating the intricate structures of medical data, thereby enhancing diagnostic and administrative capabilities within healthcare settings. Overall, this comprehensive analysis supports the potential of LLE, optimized with artificial intelligence techniques, as a pivotal tool in medical data management, offering valuable insights and methodologies to researchers and practitioners in the fields of healthcare informatics, data simplification, and pattern recognition.

## Future Directions

This paper marks the beginning of a broader exploration into applying LLE, optimized with Ai techniques, to complex medical datasets. Future research will focus on two promising areas within medical imaging. First, enhancing real-time imaging technologies such as dynamic MRI or CT scans to accelerate diagnostics and improve anomaly detection accuracy. Second, the development of automated systems using LLE to facilitate early disease diagnosis, including cancers and neurodegenerative disorders, by simplifying high-dimensional imaging data for faster and more precise analysis. These initiatives aim to leverage LLE's data simplification capabilities to advance medical diagnostics and patient care, setting the stage for significant contributions in medical imaging.

## References

[1] J. CHEN and Z. MA, "Locally linear embedding: A Review," International Journal of Pattern Recognition and Artificial Intelligence, vol. 25, no. 07, pp. 985–1008, 2011. doi:10.1142/s0218001411008993

[2] S. T. Roweis and L. K. Saul, "Nonlinear dimensionality reduction by locally linear embedding," Science, vol. 290, no. 5500, pp. 2323–2326, 2000. doi:10.1126/science.290.5500.2323

[3] B. Ghojogh, A. Ghodsi, F. Karray, and M. Crowley, "Locally Linear Embedding and its Variants: Tutorial and Survey," arXiv preprint, 2020. [Online]. Available: https://arxiv.org/abs/2011.10925

[4] "Locally Linear Embedding in R", 2009.10.05, DataMining pp. 36-350. [Online]. Available: https://www.stat.cmu.edu/~cshalizi/350/lectures/14/lecture-14.pdf

[5] X.-F. Liu, X.-D. Zlieng. G.-C. Xu, L. Wang. and H. Yang. "Locally linear embedding-based seismic attribute extraction and applications." Applied Geophysics. vol. 7. no. 4. pp. 365-375. 2010. doi:10.1007/s11770-010-0260-2

[6] X. Liu, D. Tosun, M. W. Weiner, and N. Schuff, "Locally linear embedding (LLE) for MRI based Alzheimer's Disease Classification," NeuroImage, vol. 83, pp. 148–157, 2013. doi: 10.1016/j.neuroimage.2013.06.033

[7] Y. Fang, M. Sun, S. V. N. Vishwanathan, and K. Ramani, "SLLE: Spherical locally linear embedding with applications to tomography," CVPR 2011, 2011. doi:10.1109/cvpr.2011.5995563

[8] Y. Dadmohammadi, S. Gebreyohannes, B. J. Neely, and K. A. M. Gasem, "Application of Modified NRTL Models for Binary LLE Phase Characterization," Industrial & Engineering Chemistry Research, vol. 57, no. 21, pp. 7282–7290, Apr. 2018, doi: https://doi.org/10.1021/acs.iecr.8b00683.

[9] Yanne Kouomou Chembo, D. Gomila, Mustapha Tlidi, and Curtis C.R. Menyuk, "Theory and applications of the Lugiato-Lefever Equation," The European Physical Journal D, vol. 71, no. 11, Nov. 2017, doi: https://doi.org/10.1140/epjd/e2017-80572-0.

[10] X. Wang, Y. Zheng, Z. Zhao, and J. Wang, "Bearing Fault Diagnosis Based on Statistical Locally Linear Embedding," Sensors, vol. 15, no. 7, pp. 16225–16247, Jul. 2015, doi: https://doi.org/10.3390/s150716225.

[11] G. Chen, "Dimensionality reduction of hyperspectral imagery using improved locally linear embedding," Journal of Applied Remote Sensing, vol. 1, no. 1, p. 013509, Mar. 2007, doi: https://doi.org/10.1117/1.2723663.